\documentclass[sigconf]{acmart}

\usepackage{booktabs} 

\usepackage{tikz}
\usetikzlibrary{arrows,decorations.pathmorphing,backgrounds,fit,shapes.geometric,intersections,positioning,decorations.pathreplacing,3d,circuits.logic.US}

\setcopyright{rightsretained}

\copyrightyear{2017} 
\acmYear{2017} 
\setcopyright{rightsretained} 
\acmConference{GECCO '17 Companion}{July 15-19, 2017}{Berlin, Germany}\acmDOI{http://dx.doi.org/10.1145/3067695.3075615}
\acmISBN{978-1-4503-4939-0/17/07}

\begin{document}
\title{Multitask Evolution with Cartesian Genetic Programming}
\subtitle{Extended Abstract}

\author{Eric O.\ Scott}
\affiliation{%
  \institution{George Mason University}
  \streetaddress{4400 University Drive}
  \city{Fairfax} 
  \state{Virginia} 
  \postcode{22030}
}
\email{escott8@gmu.edu}

\author{Kenneth A.\ De~Jong}
\affiliation{%
  \institution{George Mason University}
  \streetaddress{4400 University Drive}
  \city{Fairfax} 
  \state{Virginia} 
  \postcode{22030}
}
\email{kdejong@gmu.edu}


\begin{abstract}
We introduce a genetic programming method for solving multiple Boolean circuit synthesis tasks simultaneously.  This allows us to solve a set of elementary logic functions twice as easily as with a direct, single-task approach.
\end{abstract}

%
%




\maketitle

\section{Introduction}
\setlength{\belowcaptionskip}{-10pt}

Most evolutionary algorithms and metaheuristics built to date have been used as \emph{single-task} problem solvers.  With a few exceptions in areas like case-based reasoning \cite{Cunningham1997} and robot shaping \cite{DorigoColombetti}, traditionally there is no attempt to store solutions, subcomponents, or other evolved information from one task and to reuse it on other tasks.  Instead, these search algorithms rely on direct guidance from an objective function to approach a solution via variation and selection.

Very recently, a disjoint handful of researchers have begun to design new kinds of evolutionary systems that depart significantly from this single-task, adaptationist paradigm.  Preliminary efforts with \emph{multitask} evolutionary systems have begun to reveal cases where transferring and re-using information across superficially dissimilar tasks can serve to help a problem-solving agent perform more efficiently, or to provide a means of solving problems that would otherwise be deceptive or intractable.  A number of different mechanisms for effectively evolving solutions to multiple tasks simultaneously or in sequence have recently been proposed by authors working in different domains.  New structured-population models such as MAP-elites, for instance, have been developed for robotics and design applications \cite{NguyenEtAl2016}, `multifactorial' methods have been introduced for combinatorial optimization \cite{GuptaEtAl2016}, and agent-based `path-reuse' mechanisms have been proposed to allow deep neural networks that have been evolved for different pattern recognition tasks to reuse subcomponents that were trained for previous tasks \cite{FernandoEtAl2017}.

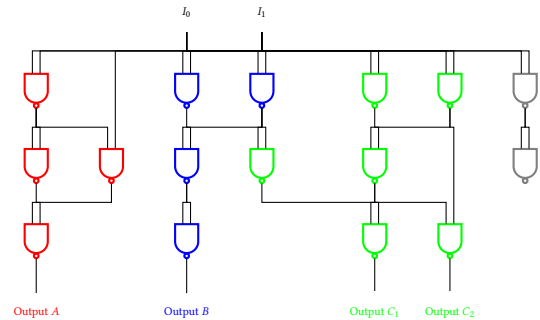
\begin{figure}
\begin{center}
\begin{tikzpicture}[circuit logic US, scale=0.5, every circuit symbol/.style={thick}]
	\node[black,buffer gate,point down,draw=none] (node0) at (0,0) {\rotatebox{90}{$I_{0}$}};
	\node[black,buffer gate,point down,draw=none] (node1) at (2,0) {\rotatebox{90}{$I_{1}$}};
	
	\node[red,nand gate,inputs={nn}, point down] (nodeA1) at (-4,-2) {};
		\draw (node0.output) -- ++(down:5mm) -| (nodeA1.input 1);
		\draw (node1.output) -- ++(down:5mm) -| (nodeA1.input 2);
	\node[red,nand gate,inputs={nn}, point down] (nodeA2) at (-2,-4) {};
		\draw (node0.output) -- ++(down:5mm) -| (nodeA2.input 1);
		\draw (nodeA1.output) -- ++(down:5mm) -| (nodeA2.input 2);
	\node[red,nand gate,inputs={nn}, point down] (nodeA3) at (-4,-4) {};
		\draw (nodeA1.output) -- ++(down:5mm) -| (nodeA3.input 1);
		\draw (nodeA1.output) -- ++(down:5mm) -| (nodeA3.input 2);
	\node[red,nand gate,inputs={nn}, point down] (nodeA4) at (-4,-6) {};
		\draw (nodeA3.output) -- ++(down:5mm) -| (nodeA4.input 1);
		\draw (nodeA2.output) -- ++(down:5mm) -| (nodeA4.input 2);
	\node[red,buffer gate, point down,draw=none] (outA) at (-4,-8) {\rotatebox{90}{Output $A$}};
		\draw (nodeA4.output) -- ++(down:5mm) -| (outA);
		
	\node[blue,nand gate,inputs={nn}, point down] (nodeB1) at (0,-2) {};
		\draw (node0.output) -- ++(down:5mm) -| (nodeB1.input 1);
		\draw (node1.output) -- ++(down:5mm) -| (nodeB1.input 2);
	\node[blue,nand gate,inputs={nn}, point down] (nodeB2) at (2,-2) {};
		\draw (node1.output) -- ++(down:5mm) -| (nodeB2.input 1);
		\draw (node0.output) -- ++(down:5mm) -| (nodeB2.input 2);
	\node[blue,nand gate,inputs={nn}, point down] (nodeB3) at (0,-4) {};
		\draw (nodeB1.output) -- ++(down:5mm) -| (nodeB3.input 1);
		\draw (nodeB2.output) -- ++(down:5mm) -| (nodeB3.input 2);
	\node[green,nand gate,inputs={nn}, point down] (nodeB4) at (2,-4) {};
		\draw (nodeB2.output) -- ++(down:5mm) -| (nodeB4.input 1);
		\draw (nodeB2.output) -- ++(down:5mm) -| (nodeB4.input 2);
	\node[blue,nand gate,inputs={nn}, point down] (nodeB5) at (0,-6) {};
		\draw (nodeB3.output) -- ++(down:5mm) -| (nodeB5.input 1);
		\draw (nodeB3.output) -- ++(down:5mm) -| (nodeB5.input 2);
	\node[blue,buffer gate, point down,draw=none] (outB) at (0,-8) {\rotatebox{90}{Output $B$}};
		\draw (nodeB5.output) -- ++(down:5mm) -| (outB);
		
	\node[green,nand gate,inputs={nn}, point down] (nodeC1) at (5,-2) {};
		\draw (node0.output) -- ++(down:5mm) -| (nodeC1.input 1);
		\draw (node1.output) -- ++(down:5mm) -| (nodeC1.input 2);
	\node[green,nand gate,inputs={nn}, point down] (nodeC2) at (7,-2) {};
		\draw (node0.output) -- ++(down:5mm) -| (nodeC2.input 1);
		\draw (node1.output) -- ++(down:5mm) -| (nodeC2.input 2);
	\node[green,nand gate,inputs={nn}, point down] (nodeC3) at (5,-4) {};
		\draw (nodeC1.output) -- ++(down:5mm) -| (nodeC3.input 1);
		\draw (nodeC2.output) -- ++(down:5mm) -| (nodeC3.input 2);
	\node[green,nand gate,inputs={nn}, point down] (nodeC4) at (5,-6) {};
		\draw (nodeB4.output) -- ++(down:5mm) -| (nodeC4.input 1);
		\draw (nodeC3.output) -- ++(down:5mm) -| (nodeC4.input 2);
	\node[green,nand gate,inputs={nn}, point down] (nodeC5) at (7,-6) {};
		\draw (nodeC2.output) -- ++(down:5mm) -| (nodeC5.input 1);
		\draw (nodeC3.output) -- ++(down:5mm) -| (nodeC5.input 2);
	\node[green,buffer gate, point down,draw=none] (outC) at (5,-8) {\rotatebox{90}{Output $C_1$}};
		\draw (nodeC4.output) -- ++(down:5mm) -| (outC);
	\node[green,buffer gate, point down,draw=none] (outC2) at (7,-8) {\rotatebox{90}{Output $C_2$}};
		\draw (nodeC5.output) -- ++(down:5mm) -| (outC2);
		
	\node[gray,nand gate,inputs={nn}, point down] (nodeD1) at (9,-2) {};
		\draw (node0.output) -- ++(down:5mm) -| (nodeD1.input 1);
		\draw (node1.output) -- ++(down:5mm) -| (nodeD1.input 2);
	\node[gray,nand gate,inputs={nn}, point down] (nodeD2) at (9,-4) {};
		\draw (nodeD1.output) -- ++(down:5mm) -| (nodeD2.input 1);
		\draw (nodeD1.output) -- ++(down:5mm) -| (nodeD2.input 2);
		
\end{tikzpicture}
\end{center}
\caption{A hypothetical logic circuit that solves three distinct Boolean functions over two inputs.  The tree that solves task \(C\) reuses subtrees of the tree that solves task \(B\).}
\label{fig:circuit}
\end{figure}

In this work, we are applying Cartesian genetic programming (CGP, \cite{Miller2011}) in a multitask fashion to evolve logic circuits that solve elementary Boolean functions.  Taking inspiration from the canonical literature on multitask learning in neural networks \cite{Caruana1998}, we follow what we call a `multi-behavior' approach to multitask evolution: each individual genome in the evolutionary algorithm encodes a solution to several tasks at once (Figure~\ref{fig:circuit}).  In the resulting program tree, the solutions to each task all share the same inputs, but have their own designated outputs.  This allows subtrees that are useful in the solution to one task to be reused as partial solutions to other tasks.

CGP has often been applied in much this way to synthesize solutions to single tasks with more than one output.  Our focus here is somewhat different: we are interested in finding ways of sharing and transferring information across multiple \emph{distinct tasks}.  The question here is whether we can use multitask CGP to more efficiently solve a set of tasks that would traditionally be treated separately.

\section{Methodology}

The set of tasks used in this study is a simple suite \(\mathcal{T}\) of 9 elementary logic function synthesis tasks that were originally used by \citeauthor{Lenski-etal} to demonstrate that complex tasks are sometimes easier to solve in conjunction with other tasks than they are to solve directly \cite{Lenski-etal}:
\begin{equation}
\mathcal{T} = \{ \texttt{AND}, \texttt{AND\_N}, \texttt{EQU}, \texttt{NAND}, \texttt{NOR}, \texttt{NOT}, \texttt{OR}, \texttt{OR\_N}, \texttt{XOR} \}. \nonumber
\end{equation}
Our primitive set consists only of \(\{ \texttt{NAND} \}\). 

For a preliminary proof of concept we wish only to demonstrate the following claim:\\[-1.5ex]

\hangindent=0.4cm
\textbf{Hypothesis}: A multi-behavior approach will be able to solve all \(|\mathcal{T}| = 9\) tasks using less computational effort than a traditional single-task approach.\\[-1.5ex]

\noindent To test this, we apply a standard \((1+4)\)-style implementation of Cartesian genetic programming to a fitness function that evaluates each circuit on all 9 tasks simultaneously.  We treat the average fitness across all 9 tasks as a scalar fitness value.

Now, in multitask algorithm applications it sometimes happens that progress is impeded by \emph{task interference}: progress on one task may overwrite the solution to previously solved tasks, preventing the algorithm from encoding solutions to many tasks at once.  In light of this, we further hypothesize that multi-behavior CGP will perform better if we avoid mutating genes that belong to a successful solution to a task.  We effect this by tracking which outputs each circuit element contributes information to.  We then configure the probability that each element's genes are mutated to be a decreasing function of the mean fitness of all the tasks it belongs to.  We test this weighted mutation scheme with both linearly- and exponentially-decreasing weighting functions. All together, we implemented one single-task CGP algorithm and three multi-behavior algorithms (a constant-mutation case, the linear case, and the exponential case).

In order to achieve a fair comparison, we performed an extensive sweep over CGP's free parameters and our weighted mutation scheme parameters to select configurations of each algorithm that solve the suite of tasks in the least computational effort on average.  As a result of this parameter-tuning process, some algorithms are configured to use different circuit sizes than others.  To compare results with different circuit sizes, our measure of computational effort is \emph{node-evaluations}---that is, the number of individuals that have been evaluated, multiplied by their size. To solve all 9 tasks with each algorithm, we run each multi-behavior algorithm once on the task suite, and then we run the single-task algorithm 9 times in sequence, once for each of the tasks, measuring the total number of node-evaluations expended along the way before a solution is found.

\section{Preliminary Results}

\begin{figure}
\includegraphics[width=0.9\columnwidth]{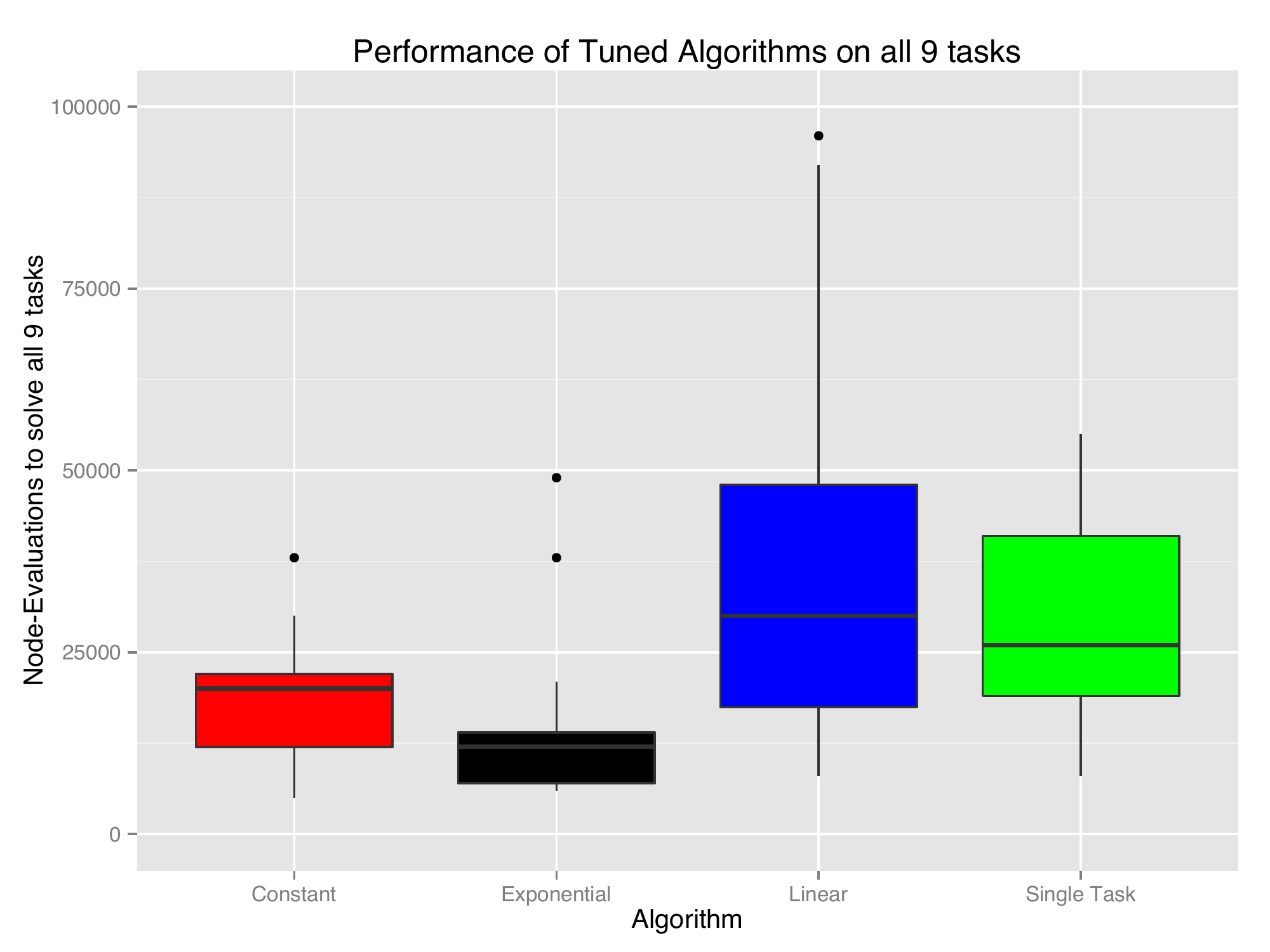}
\caption{Preliminary results indicate that a multi-behavior approach is more efficient at solving all 9 tasks than a single-task approach.}
\label{fig:results}
\end{figure}

Results on the 9-logic suite suggest that the multi-behavior CGP algorithm is more efficient than a single-task approach when a constant or exponentially-weighted mutation rate is used.  Figure~\ref{fig:results} shows the distribution of effort each algorithm required to successfully find solutions to all 9 tasks. On average, we find that the exponentially-weighted algorithm solves the tasks twice as easily as the single-task algorithm does.

One of the potential advantages of this kind of multitask evolution is that no prior knowledge is needed about how information might be reused across different tasks: evolution dynamically discovers subproblems that are redundant across tasks.  Much more work is needed, however, to understand how sharing information across tasks works in different domains, and what conditions or structural similarities among tasks are necessary to see performance improvements from multitask evolution.

As this research continues, we intend to analyze the circuits generated by the multi-behavior approach for patterns of reuse, and to extend this methodology to more complex task suites and primitive sets.

\bibliographystyle{ACM-Reference-Format}
\bibliography{sigproc} 

\end{document}